\documentclass[lettersize,journal]{IEEEtran}
\usepackage{amsmath,amsfonts}
\usepackage{algorithmic}
\usepackage{algorithm}
\usepackage{array}
\usepackage[caption=false,font=normalsize,labelfont=sf,textfont=sf]{subfig}
\usepackage{textcomp}
\usepackage{stfloats}
\usepackage{url}
\usepackage{verbatim}
\usepackage{graphicx}
\usepackage{cite}
\usepackage{multicol}
\usepackage{multirow}
\usepackage{booktabs}

\usepackage{color}
\begin{document}

\title{Biomedical Relation Extraction via Adaptive Document-Relation Cross-Mapping and Concept Unique Identifier}

\author{Yufei Shang,
    Yanrong Guo\textsuperscript{*},
    Shijie Hao,
    and Richang Hong\textsuperscript{*}
    \thanks{Yufei Shang, Yanrong Guo, Shijie Hao, and Richang Hong are with the School of Computer Science and Information Engineering, Hefei University of Technology, Hefei 230009, China. Yanrong Guo and Richang Hong are the corresponding authors (yrguo@hfut.edu.cn, hongrc.hfut@gmail.com).}}

\maketitle

\begin{abstract}
Document-Level Biomedical Relation Extraction (Bio-RE) aims to identify relations between biomedical entities within extensive texts, serving as a crucial subfield of biomedical text mining.
Existing Bio-RE methods struggle with cross-sentence inference, which is essential for capturing relations spanning multiple sentences. Moreover, previous methods often overlook the incompleteness of documents and lack the integration of external knowledge, limiting contextual richness. Besides, the scarcity of annotated data further hampers model training.
Recent advancements in large language models (LLMs) have inspired us to explore all the above issues for document-level Bio-RE.
Specifically, we propose a document-level Bio-RE framework via LLM Adaptive Document-Relation Cross-Mapping (ADRCM) Fine-Tuning and Concept Unique Identifier (CUI) Retrieval-Augmented Generation (RAG).
First, we introduce the Iteration-of-REsummary (IoRs) prompt for solving the data scarcity issue. In this way, Bio-RE task-specific synthetic data can be generated by guiding ChatGPT to focus on entity relations and iteratively refining synthetic data.
Next, we propose ADRCM fine-tuning, a novel fine-tuning recipe that establishes mappings across different documents and relations, enhancing the model's contextual understanding and cross-sentence inference capabilities.
Finally, during the inference, a biomedical-specific RAG approach, named CUI RAG, is designed to leverage CUIs as indexes for entities, narrowing the retrieval scope and enriching the relevant document contexts.
Experiments conducted on three Bio-RE datasets—GDA, CDR, and BioRED—demonstrate the state-of-the-art performance of our proposed method by comparing it with other related works. 
\end{abstract}


\section{Introduction}
\IEEEPARstart{B}{iomedical} Relation Extraction (Bio-RE) plays a crucial role in the field of biomedical text mining, aiming to identify the relations between two entities within biomedical texts automatically. Bio-RE is pivotal in developing applications such as medical knowledge graph construction, question-answering systems, and biomedical text analysis, which enhances the accessibility and comprehension of complex biological data.

Generally, Bio-RE is classified into two primary categories based on the length of text processed: document-level and sentence-level.
Prior studies primarily concentrate on sentence-level RE \cite{Distantly-Supervised-RE,FPrompt-PLM,KnowPrompt,Biomedical_Distant_SupervisionRE, REBEL, LearningRelationPrototype}.
However, in real-world scenarios, a significant number of relational facts are expressed across multiple sentences.
Research indicates that over 40.7\% of relational facts necessitate the analysis of multiple sentences \cite{yao-etal-2019-docred}, illustrating the complexity and value involved in document-level Bio-RE.

As for document-level RE, a considerable amount of information regarding entities and their relations within a document can only be identified through cross-sentence analysis\cite{TKDEDocRE}.  
The need for cross-sentence inference is especially critical in biomedical documents. Unlike general-domain texts, biomedical documents often contain aliases and identical terms sometimes exhibiting polysemy, thereby referring to entirely different entities.
Moreover, the high level of professionalism and logical structure in biomedical texts intensifies the demand for robust cross-sentence inference in document-level Bio-RE compared to conventional document-level RE.
  
\begin{figure}[!t]
\centering
\includegraphics[width=0.9\columnwidth]{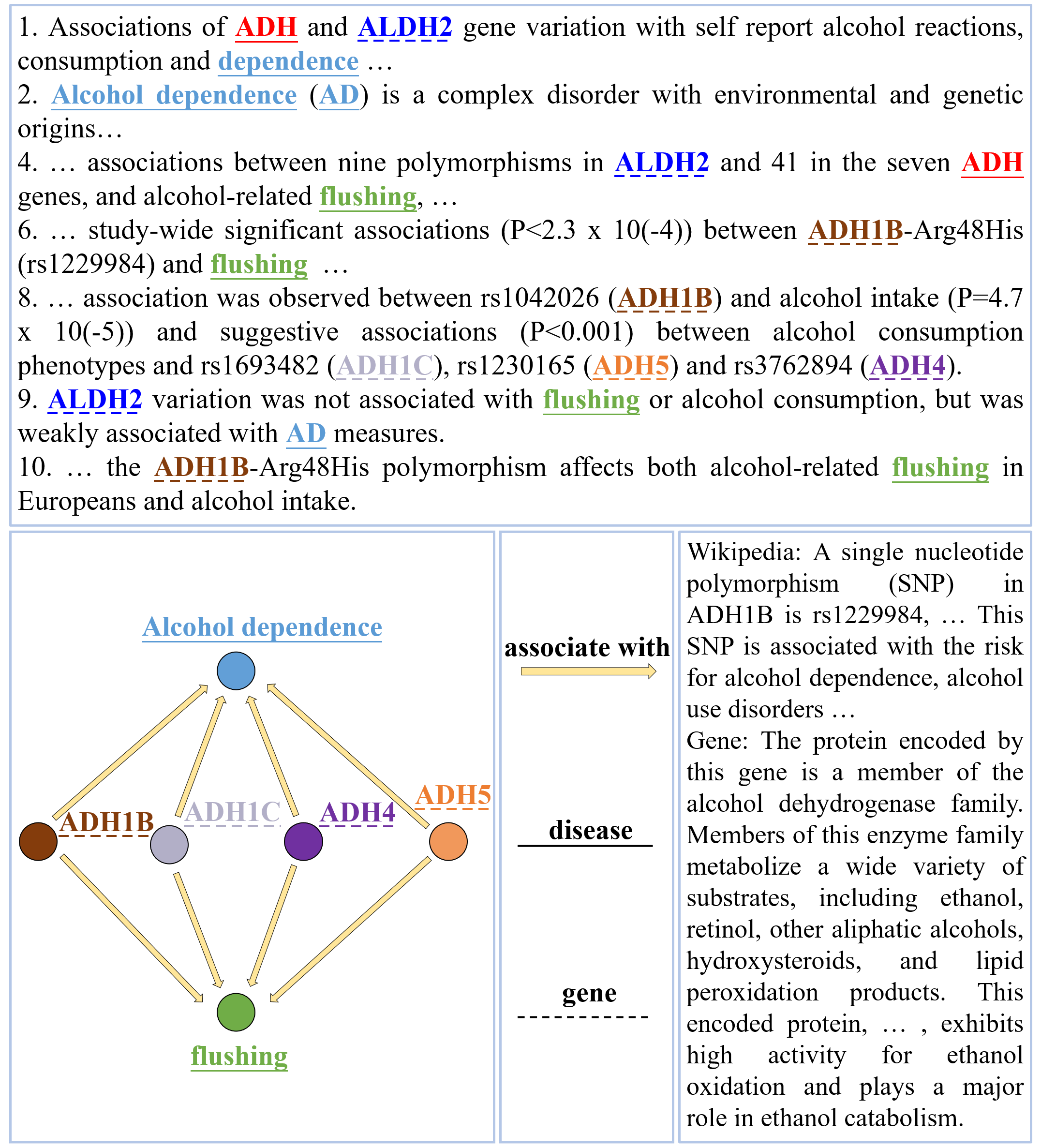}
\caption{This figure illustrates a document-level Bio-RE example from the GDA dataset \cite{10.1007/GDA}. Mentions of the same entity are highlighted in consistent colors for clarity. Solid underlines indicate disease entities, while dashed underlines represent gene entities. The lower right corner shows the retrieval results for the ADH1B gene from Wikipedia and National Center for Biotechnology Information Gene database.}
\label{fig_1}
\end{figure}
As illustrated in Figure \ref{fig_1}, the Gene Disease Association (GDA) between the \textit{ADH1B} gene and \textit{flushing} is an intra-sentence relation, explicitly stated in sentence six. However, the other seven GDAs identified are inter-sentence relations, highlighting the importance of cross-sentence capabilities in RE. 
Additionally, the entity \textit{alcohol dependence} appears in sentences 1, 2, and 9, and is referred to by aliases such as \textit{dependence} and \textit{AD}. This exemplifies the critical need for cross-sentence inference within document-level RE to effectively integrate information from multiple mentions of the same entity, presented under various aliases and forms.

Furthermore, documents for document-level RE often come from sources like Wikipedia and Wikidata \cite{Wikidata}, which provide detailed explanations. In contrast, documents for document-level Bio-RE are typically sourced from more condensed materials, such as biomedical article abstracts, which may lack comprehensive information.
Consequently, a significant challenge faced by document-level Bio-RE is the increasing necessity to draw upon external world knowledge. This need arises not only to compensate for the inherent incompleteness of documents but also to provide more accurate and referable contexts. For instance, in Figure \ref{fig_1}, by retrieving information on \textit{ADH1B} from sources such as Wikipedia and National Center for Biotechnology Information (NCBI), we learn that a single nucleotide polymorphism in \textit{ADH1B} is associated with the risk for \textit{alcohol dependence} and that \textit{ADH1B} exhibits high activity in the oxidation process of ethanol.

Another major challenge in document-level Bio-RE is the scarcity of annotated data.
For example, the CDR dataset \cite{10.1093/database/CDR}, a key resource for chemical-disease RE, includes only 500 documents in its training set. This is considerably fewer than the general-domain DocRED dataset\cite{yao-etal-2019-docred}, which provides 104,926 documents in its training set.
Not only is the amount of annotated data limited, but the inherent professionalism and logical structure of biomedical documents also make the manual annotation process time-consuming, labor-intensive, and highly specialized.
The shortage of well-annotated data hampers the development and refinement of Bio-RE models that predominantly rely on large datasets for training and validation.

\begin{figure}[!t]
\centering
\includegraphics[width=0.9\columnwidth]{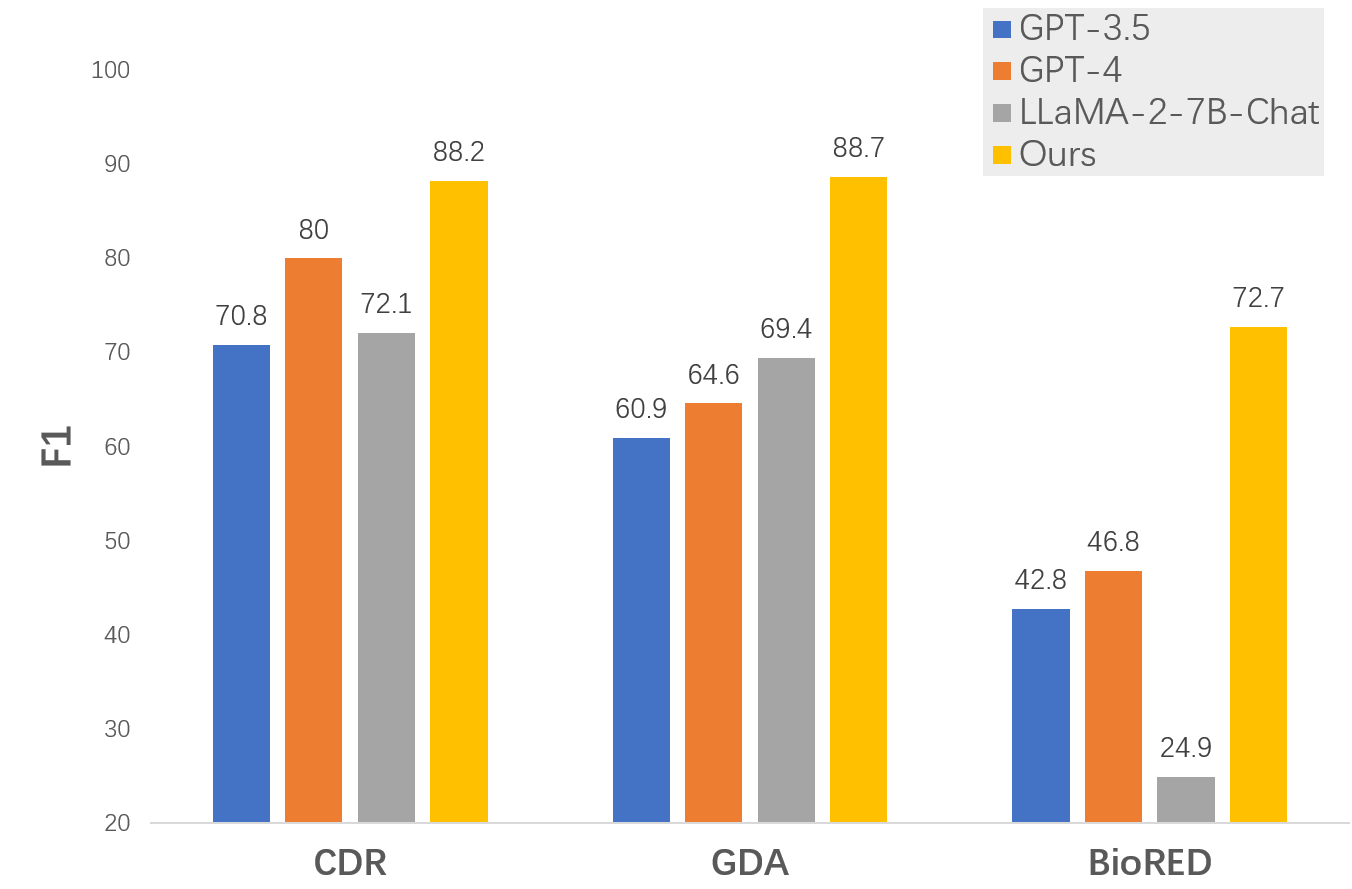}
\caption{The performance of LLMs on the test sets of the CDR, GDA, and BioRED datasets.}
\label{fig_LLM}
\end{figure}
With the recent development of LLMs such as ChatGPT \cite{openai2024gpt4} and LLaMA2 \cite{LLaMA2}, there has been growing research interest in leveraging LLMs for document-level RE \cite{wei2024chatie,wan2023gpt,Xue_AutoRE}. Consequently, we evaluated several LLMs on the document-level Bio-RE task, using test sets from the CDR \cite{10.1093/database/CDR}, GDA \cite{10.1007/GDA}, and BioRED \cite{BioRED} datasets. As shown in Figure \ref{fig_LLM}, the models evaluated include GPT-3.5, GPT-4, and LLaMA2-7B-Chat. Our results indicate that the direct application of LLMs to the document-level Bio-RE task yields suboptimal performance, particularly on the BioRED dataset, which involves a multi-class classification scenario.
LLMs face significant limitations when directly applied to the document-level Bio-RE task, as they lack the necessary medical knowledge and effective fine-tuning specifically for document-level Bio-RE. Furthermore, when faced with complex, incomplete, and cross-sentence inference-intensive biomedical documents, their sophisticated text analysis capabilities fall short.

To address the aforementioned challenges of document-level Bio-RE and the limitations of directly applying LLMs, we introduce a novel framework for document-level Bio-RE via LLM Adaptive Document-Relation Cross-Mapping (ADRCM) fine-tuning and Concept Unique Identifier (CUI) Retrieval-Augmented Generation (RAG), specifically designed to enhance document-level Bio-RE. We evaluated this framework on three document-level Bio-RE datasets: GDA \cite{10.1007/GDA}, CDR \cite{10.1093/database/CDR}, and BioRED \cite{BioRED}, where it achieves state-of-the-art performance across all. Our contributions are summarized as follows:
\begin{itemize}
\item{
We propose ADRCM fine-tuning, a novel fine-tuning recipe for LLMs in document-level Bio-RE, which establishes mutual mappings between documents and relations, enabling the model to capture domain-specific language nuances and enhance cross-sentence inference.
}
\item{We develop CUI RAG, which uses CUIs as indexes for entities, not only narrowing the retrieval scope and enhancing relevance in specialized biomedical contexts, but also reducing the impact of different aliases for entities on retrieval.}
\item{We propose the Iteration-of-REsummary (IoRs) prompt, which guides ChatGPT to generate focused summaries by concentrating on specified entity relations and iteratively refining the data. This cost-effective strategy enhances the generalization and accuracy of LLM without significantly increasing annotation costs.
}
\end{itemize}

\section{Related Work}
Current methods for document-level RE including document-level Bio-RE, can be primarily categorized into graph-based, transformer-based, and LLM-based methods.

\textbf{Graph-based Methods.} These methods typically build a document-level graph using words, mentions, entities, or sentences as nodes, and predict relations by performing reasoning on the graph. 
Christopoulou et al.\cite{christopoulou-etal-2019-connecting} proposed an edge-oriented model for document-level relation extraction that emphasizes edge representations over node representations to more effectively model entity relations. The model constructs nodes at various levels, including sentence, mention, and entity levels, and employs a partially-connected document graph with heterogeneous node and edge types. 
LSR \cite{Nan_Guo_Sekulic_Lu_2020} treats the graph structure as a latent variable, automatically inducing the optimal structure in an end-to-end manner without relying on pre-defined syntactic or co-reference structures. It employs an iterative refinement strategy that incrementally improves the latent structure, enabling the model to dynamically refine the graph across multiple iterations for effective multi-hop reasoning. 

To further enhance relational reasoning over graphs, several efforts were made to design specialized reasoning networks \cite{peng2022document,Inter_Pair_Reasoning2023,Document_Level_Relation_Extraction_Path_Reasoning2023,Reconstruction2021}.
For example, SGR \cite{peng2022document} focuses on extracting a simplified subgraph around the target entity pair, which contains the most relevant paths for relational reasoning. The approach generates reasoning paths through a heuristic strategy that explicitly models essential reasoning skills, such as logical reasoning and co-reference resolution. By applying a Relational Graph Convolutional Network to the extracted subgraph, SGR allows the model to focus on the most crucial entities, mentions, and sentences, enabling more effective joint reasoning over multiple paths.
Xu et al. \cite{Document_Level_Relation_Extraction_Path_Reasoning2023} proposed a novel path reasoning method that uses a breadth-first search (BFS) algorithm to extract multiple reasoning paths in a document-level graph. The extracted paths are then encoded using a long-short term memory (LSTM) network, and an attention layer is employed to summarize these paths, simulating complete reasoning paths between entities.
To better differentiate the importance of various nodes and edges while filtering out irrelevant information, several studies integrated attention mechanisms into these models \cite{DAGCN,tian-etal-2021-dependency,hang2024graphAtt}. For example, DAGCN \cite{DAGCN} establishes bidirectional information flow and enables multi-turn interactions between contextual and dependency information through a parallel structure. Additionally, it employs a multi-layer Adjacency Matrix-Aware Multi-Head Attention mechanism, which effectively preserves the structural information of sentences and dependency trees during interactions.

Graph-based document-level Bio-RE methods share similarities with general graph-based document-level RE methods in their underlying approach. For example, both Topic-BiGRU-U-Net \cite{Topic-BiGRU-U-Net} and FILR \cite{li-etal-2022-document} integrate contextual information with graph-based representations and employ specialized multi-granularity reasoning networks to capture interactions between entities and mentions across sentences in biomedical texts. Additionally, AGCN \cite{AGCN}, DAM-GAN \cite{Dual-Attention_bioRE} and HTGRS \cite{HTGRS} incorporate attention mechanisms into graph-based methods to enhance their effectiveness.

However, graph-based methods are significantly influenced by the quality of the constructed document-level graph and typically consider only edge and entity information during relational reasoning. 
These methods often neglect many non-entity clues present in the document, thereby limiting the further enhancement of the model’s reasoning capability.

\textbf{Transformer-based Methods.}
Transformer-based methods leverage the capability of pre-trained language models (PLMs) to capture long-range dependencies by implicitly modeling long-distance relations through multi-head attention. These methods have gained significant attention for their effectiveness in performing relational reasoning and enhancing entity representations.
SSAN \cite{xu2021Entitystructurewithinandthroughout} uses a unified framework to capture various mention dependencies and fully integrates structural dependencies within the encoding network. It extends the self-attention mechanism by incorporating Biaffine and Decomposed Linear Transformations, allowing the model to capture entity relations and structural dependencies across the document more effectively.
ALOTP \cite{zhou2021documentALOTP} employs adaptive thresholding, replacing the traditional global threshold with a learnable, entity pair-specific threshold that allows the model to dynamically adjust to different entity pairs. Moreover, it utilizes localized context pooling, refining entity embeddings by focusing on the context most relevant to each specific entity pair.
DocRE-II \cite{zhang-etal-2022-IterativeInference} initially predicts relations and then iteratively refines them using Extended Cross Attention units, which capture dependencies among overlapping entity pairs by integrating both feature-level and relation-level information.
SAIS \cite{xiao-etal-2022-sais} enhances RE by explicitly supervising intermediate steps through four tasks: Coreference Resolution, Entity Typing, Pooled Evidence Retrieval, and Fine-grained Evidence Retrieval. These tasks help the model capture textual contexts and entity types more effectively, leading to more accurate and interpretable RE. Additionally, SAIS employs evidence-based data augmentation, selectively refining predictions when model uncertainty is detected. DocRE-SD \cite{zhang2023self-distillation} introduces a reasoning multi-head self-attention mechanism that models four common reasoning patterns, enhancing relational triple coverage. It also employs a self-distillation framework to explicitly model relational reasoning by masking entity pairs during training. Additionally, a curriculum learning strategy is used to gradually increase the complexity of masked pairs, resulting in more robust learning.

Building on the success of Transformer-based methods in document-level RE, similar strategies were adopted in the realm of document-level Bio-RE. For instance, TriA-BioRE \cite{TriA-BioRE}, incorporating a Triangular Attention Module, enhances pair-level modeling for Bio-RE by comprehensively capturing interdependencies between entity pairs through a combination of triangular multiplications and self-attention mechanisms.

Transformer-based methods, although powerful, have limitations in document-level RE due to a fixed maximum input length, which restricts their ability to effectively handle long documents by potentially truncating important contexts. Additionally, in specific domains, PLMs often struggle to keep pace with the latest knowledge. Continuous pretraining or fine-tuning is necessary to keep PLMs updated with the most recent information. However, this process is resource-intensive and demands access to up-to-date, high-quality training data. 

\textbf{LLM-based Methods.}
In recent years, the rise of LLMs has revolutionized document-level Bio-RE by leveraging their vast contextual understanding, extensive pre-trained knowledge, and ability to capture complex dependencies across long textual spans. 
ChatIE \cite{wei2024chatie} leverages a multi-turn question-answering approach with ChatGPT to decompose complex information extraction (IE) tasks into simpler sub-tasks.
GPT-RE \cite{wan2023gpt}  enhances in-context learning for RE by employing task-aware demonstration retrieval and gold label-induced reasoning.
AutoRE \cite{Xue_AutoRE} employs a novel Relation-Head-Facts paradigm and Parameter Efficient Fine-Tuning (PEFT) with LLM, achieving state-of-the-art results on the Re-DocRED dataset.
Multi-Span \cite{LLMDOCRE} redefines document-level RE as a machine reading comprehension problem by transforming the identification of entities and relations into a structured question-answering process. To generate example answers, the approach integrates LLMs during the question construction phase, enhancing the model's contextual understanding and reasoning capabilities. Furthermore, it introduces a hybrid pointer-sequence labeling model that effectively handles the extraction of zero or multiple answers.
Additionally, several studies have explored the application of LLMs for code generation (Code-LLMs) in IE tasks \cite{guo2023code-generationUIE, bi2024codekgc, li2023codeie}. These methods highlight the potential of LLMs in RE by leveraging contextual understanding and pre-trained knowledge.

To the best of our knowledge, none of these approaches has introduced a fine-tuning recipe specifically tailored to document-level Bio-RE, nor have they developed a dedicated RAG approach for it, let alone integrated the two.

In summary, graph-based methods depend on the quality of constructed graphs and often overlook non-entity clues. 
Transformer-based methods are limited by a fixed maximum input length, which truncates important context.
LLM-based methods struggle to explore effective, targeted fine-tuning recipes and RAG approaches, as well as to address the scarcity of annotated document-level Bio-RE data.
Our proposed framework effectively addresses these challenges. We leverage the strong text comprehension capabilities of LLMs and introduce ADRCM fine-tuning, specifically designed for document-level Bio-RE to improve the model's cross-sentence inference abilities. This framework eliminates the need for graph structures and enables the processing of longer texts without sacrificing context. 
Additionally, we supplement the model's training data with high-quality synthetic data generated by ChatGPT and incorporate CUI RAG to provide more comprehensive and relevant document contexts, ensuring up-to-date Bio-RE.
\begin{figure*}[!t]
\centering
\includegraphics[width=0.95\textwidth]{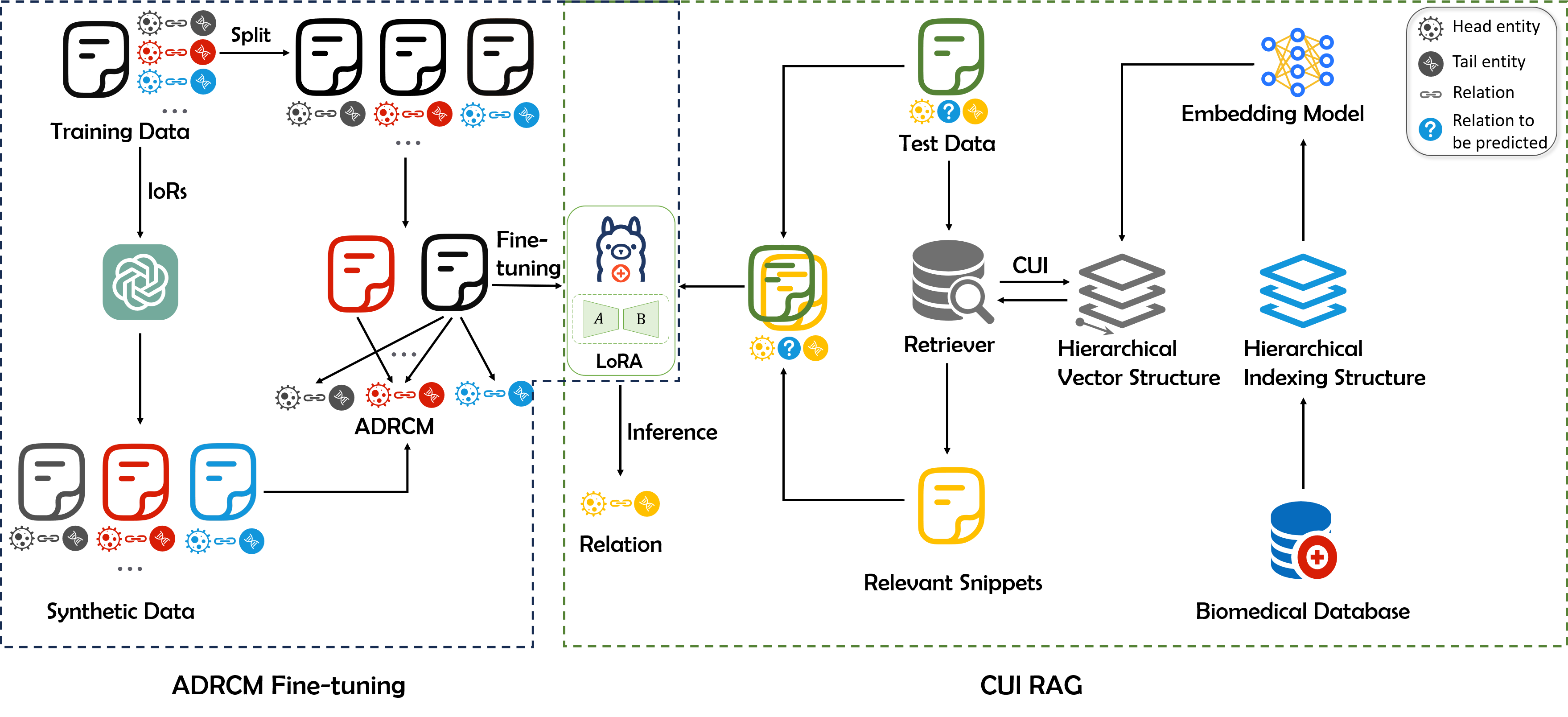}
\caption{Overview of our framework.  Gray, red, and blue are used to distinguish different entities, relations, and documents.}
\label{fig_2}
\end{figure*}
\section{Methodology}
In this section, we provide a detailed introduction to our proposed framework. As illustrated in Figure \ref{fig_2}, our framework consists of two stages: the ADRCM fine-tuning stage and the CUI RAG stage. 
In the ADRCM fine-tuning stage, the IoRs prompt iteratively guides ChatGPT to generate synthetic data labeled consistently with the original training data. The training data is split according to the head-relation-tail triplet and then merged with the synthetic data to form ADRCM-structured data. This combined data is used to fine-tune the LLaMA2-7B-Chat model with Low-Rank Adaptation (LoRA) \cite{Lora}.
In the CUI RAG stage, relevant snippets are retrieved from biomedical databases based on the entities in the test data and their corresponding CUIs. These snippets, together with the test data, are analyzed by the fine-tuned LLaMA2-7B-Chat model to determine the predicted relations.
\subsection{Task Definition}
Given a biomedical document $d_i$ containing a set of biomedical entities $\mathbb{E}_i$, with $h_{i,j}\in\mathbb{E}_i$ and
$t_{i,j}\in\mathbb{E}_i$ denoting the pair of head and tail entity. Given a predefined set of relation classes $\mathbb{R}$, the document-level Bio-RE task is to predict the relation $r_{i,j}\in\mathbb{R}$ between the pair of entities $h_{i,j},t_{i,j}$. Here, $i$ indexes the document, and $j$ indexes the entity pair within that document.
\begin{algorithm}[!t]
    \caption{Procedure  of generating synthetic data through IoRs prompt 
    }
    \label{alg:IoRs}
    \renewcommand{\algorithmicrequire}{\textbf{Input:}}
    \renewcommand{\algorithmicensure}{\textbf{Output:}}
    \begin{algorithmic}[1]
        \REQUIRE document $d_i$, head entity $h_{i,j}$, tail entity $t_{i,j}$, relation $r$, threshold $\beta$  
        \ENSURE  synthetic data $(ds_{i,j},h_{i,j},t_{i,j},r_{i,j})$ or NULL 
        \STATE  Initialize an empty list $S$ to store the summaries generated by ChatGPT
        \STATE  Initialize the summary prompt $P_s=I_s+h_{i,j}+t_{i,j}+r_{i,j}$, where $I_s$ represents the instruction for generating the summary
        \STATE  Initialize the relation confirmation prompt $P_c=I_c+h_{i,j}+t_{i,j}+r_{i,j}$, where $I_c$ represents the instruction for relation confirmation
        \STATE  Initialize a counter $\theta$ to track the number of iterations
        
        \WHILE{$\theta<\beta$}
            \STATE Generate a summary $ds_{i,j} = \text{ChatGPT}(P_s, d_i,S)$
            \STATE Perform relation confirmation to obtain confirmed relation $\Dot{r}_{i,j} = \text{ChatGPT}(P_c,ds_{i,j})$
            \IF{$\Dot{r}_{i,j} == r_{i,j}$}
                
                \RETURN $(ds_{i,j},h_{i,j},t_{i,j},r_{i,j})$
            \ELSE
                \STATE Append $ds_{i,j}$ to $S$
                \STATE  $\theta\gets \theta + 1$ 
            \ENDIF
            
        \ENDWHILE
        \RETURN NULL
    \end{algorithmic}
\end{algorithm}
\subsection{Iteration-of-REsummary (IoRs) Prompt}
\begin{figure*}[!t]
\centering
\includegraphics[width=0.9\textwidth]{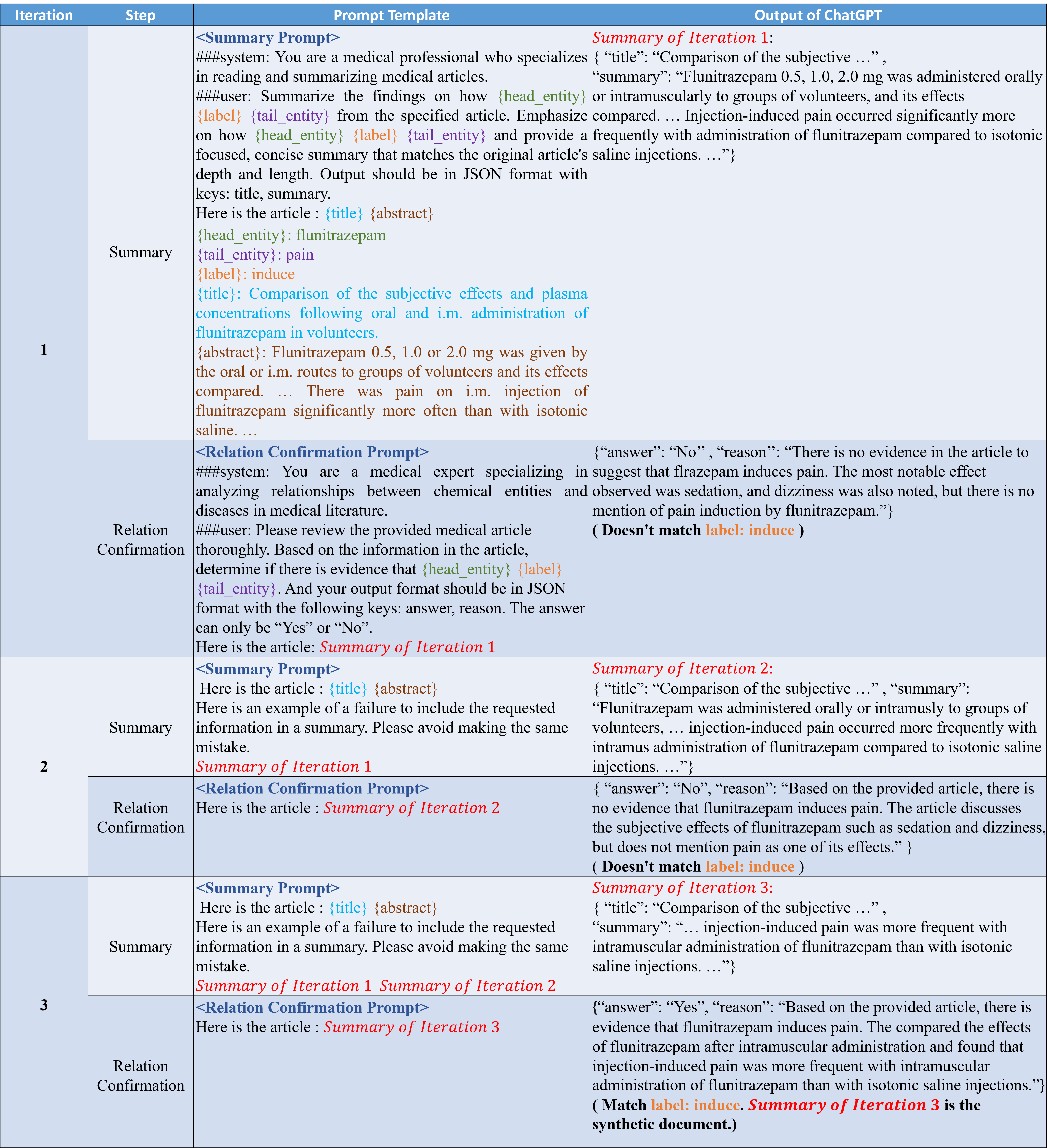}
\caption{An example of IoRs. The generation process is independent, meaning that each step does not retain the memory of the previous steps.}
\label{fig_3}
\end{figure*}
In recent years,  LLMs have been extensively applied to data augmentation due to their powerful generative capabilities. These models can produce high-quality synthetic data that closely mirrors real-world data, which is particularly valuable in the field of document-level Bio-RE. This field often faces challenges with the scarcity and high cost of obtaining annotated data, making LLMs an essential tool for improving data availability and model performance.
However, most current approaches are designed for sentence-level data \cite{xu2023_S2ynRE,ding2023gptDataAnnotator,chia2022relationprompt,LLM_based_Augmentation_2024}, which typically feature a single semantic structure, making them overly simplistic and lacking the broader context found in longer texts.
Furthermore, approaches targeting document-level data often rely on LLMs to generate multiple labels simultaneously \cite{synthetic_data1,synthetic_data2}. This practice exacerbates the issue of hallucinations, thereby reducing the reliability of the generated annotations.

To address these issues, we propose the IoRs prompt, which guides ChatGPT to summarize a specific pair of entities and their relation, ensuring that the synthetic data matches the original training labels through iteration, as illustrated in Algorithm \ref{alg:IoRs}. An example of the IoRs prompt is presented in Figure \ref{fig_3}, and the procedure for generating synthetic data is described as follows:
\begin{enumerate}
\item
We prompt ChatGPT to create a summary based on document $d_i$, the head entity $h_{i,j}$, the tail entity $t_{i,j}$, and the relation $r_{i,j}$. The prompt guides the model to focus on $h_{i,j}$, $t_{i,j}$, and $r_{i,j}$ to produce a focused and stylistically consistent summary of $d_i$, yielding the summary $ds_{i,j}$. 
\item 
Subsequently, ChatGPT is used to perform relation confirmation based on the summary $ds_{i,j}$,  head entity $h_{i,j}$, and tail entity $t_{i,j}$, to obtain the confirmed relation $\Dot{r}_{i,j}$.
\item 
Determine whether the confirmed relation $\Dot{r}_{i,j}$ matches the true relation $r_{i,j}$. If they match, then $ds_{i,j}$ is utilized as the synthetic document for this training instance, with $(ds_{i,j},h_{i,j},t_{i,j},r_{i,j})$ incorporated as a sample into the synthetic dataset.
If they do not match, $ds_{i,j}$ is treated as a failure example while keeping $d_i$, $h_{i,j}$, $t_{i,j}$, and $r_{i,j}$ unchanged, and the process returns to step 1) for further iterations.

\item
If the number of iterations exceeds a threshold $\beta$ and $\Dot{r}_{i,j}$ still does not match $r_{i,j}$, the loop is terminated and the synthetic data $(ds_{i,j},h_{i,j},t_{i,j},r_{i,j})$ is discarded for this training instance.
\end{enumerate}

\subsection{ADRCM Fine-tuning}
To enhance cross-sentence inference, contextual understanding, and focus on critical document segments for LLMs in document-level Bio-RE, we propose ADRCM fine-tuning.
This fine-tuning recipe not only leverages both the original training dataset and synthetic dataset but also establishes mappings between documents and relations, forming an Adaptive Document-Relation Cross-Mapping that enables the model to learn domain-specific language nuances and better capture complex relations across sentences.

Firstly, for each sample $o_i$ in the original training dataset $D_o$, we split it based on the triplets to create $sp_i$, in which each document corresponds to a single triplet. This process is illustrated in the following equations.
\begin{equation}
D_o = \{o_i \mid i = 1, 2, \ldots, N\}
\label{Do}
\end{equation}
\begin{equation}
    o_i=(d_i,\{(h_{i,j},t_{i,j},r_{i,j})\mid j=1,2,\ldots,J_i\})
\label{oi}
\end{equation}
\begin{equation}
o_i\xrightarrow{\mathrm{split}}sp_i=\{(d_i,h_{i,j},t_{i,j},r_{i,j})\mid j=1,2,\ldots,J_i\}
\label{split}
\end{equation}
In Equation \ref{Do}, the original training dataset $D_o$ is defined as containing $N$ samples $o_i$. Each sample $o_i$, as shown in Equation \ref{oi}, consists of a document $d_i$ and $J_i$  triplets $(h_{i,j},t_{i,j},r_{i,j})$.  In Equation \ref{split}, each $o_i$  is split into $sp_i$, a set containing $J_i$ elements, where each element is composed of the same document $d_i$ and a different triplet $(h_{i,j},t_{i,j},r_{i,j})$. This structure in $sp_i$ represents a mapping of multiple triplets to a single document.

Next, we generate synthetic data $sd_i$ corresponding to $o_i$ using the IoRs prompt.
\begin{equation}
o_i\xrightarrow{\mathrm{IoRs}}sd_i=\{(ds_{i,j},h_{i,j},t_{i,j},r_{i,j})\mid j=1,2,\ldots,J_i\}
\label{IoRs}
\end{equation}
Here, the IoRs prompt generates a different document $ds_{i,j}$  for each triplet $(h_{i,j},t_{i,j},r_{i,j})$, resulting in $sd_i$ as a mapping of each unique triplet to a distinct document.

Then, $sp_i$ is merged with the synthetic data $sd_i$ to create ADRCM-structured data $Asd_i$, which can be expressed as:
\begin{equation}
\begin{aligned}
Asd_i &= sp_i \cup sd_i \\
      &= \{(d, h_{i,j}, t_{i,j}, r_{i,j}) \mid d \in \{d_i, ds_{i,j}\},\\
      &\qquad \qquad \qquad \qquad \quad j = 1, 2, \ldots, J_i\}
\end{aligned}
\end{equation}
In this structure, $d$ represents either the original document $d_i$ or the synthetic document $ds_{i,j}$, each corresponding to a triplet $(h_{i,j},t_{i,j},r_{i,j})$. 

By iterating over all samples $o_i$ in the original training dataset $D_o$, we combine $Asd_i$ to form the ADRCM-structured dataset $AsD$. This can be expressed as follows:
\begin{equation}
AsD = \{Asd_i \mid i = 1, 2, \ldots, N\}
\end{equation}

ADRCM enables $AsD$  to include diverse entity pairs and relations mapped to the same document. Fine-tuning with this data implicitly trains the model to focus on document sections that are crucial for accurately understanding specific relations. This targeted learning process allows the model to more effectively isolate relevant information during inference.
Moreover, $AsD$ includes instances where the same entity pair and relation are mapped to different documents. Fine-tuning with such data exposes the model to varied contexts for each entity-relation pair, allowing it to develop a deeper understanding of how relational meaning shifts depending on context. This capability is particularly valuable for capturing the domain-specific nuances of biomedical texts.
Together, these characteristics foster the development of cross-sentence inference skills, enabling the model to track relational cues across different sections of a document and effectively interpret the diverse expressions of relations across sentences. This approach enhances the model's ability to capture complex, cross-sentence relations, which is essential for effective document-level Bio-RE.

Finally, we use $AsD$ to fine-tune the LLaMA2-7B-Chat model through LoRA. The fine-tuning procedure can be formally described as follows:
\begin{equation}
\widetilde{M} \leftarrow \text{LoRA}(M,I,AsD)
\end{equation}
where $\widetilde{M}$ represents the fine-tuned model obtained from the backbone model $M$. $I$ denotes the task instruction of document-level Bio-RE. 

\subsection{CUI RAG}
To address the prevalent challenges of factual hallucination \cite{zhanghallucination}, knowledge obsolescence \cite{outdate}, the lack of domain-specific knowledge in LLMs \cite{domainknowledge}, as well as the effects of biomedical entity synonymy and aliases on retrieval accuracy, we propose a specialized RAG method tailored for the biomedical domain, termed CUI RAG. This method employs Concept Unique Identifiers (CUIs) from the Unified Medical Language System (UMLS) \cite{umls} as indexes to define the retrieval scope and enhance the relevance of retrieval results. In the following sections, we provide a detailed description of our CUI RAG method. 

\begin{itemize}
    \item \textbf{Retrieval Source.} 
    We primarily use Wikipedia and several NCBI biomedical databases, such as Gene, MeSH, and Protein, as our retrieval sources. 
    These biomedical-specific sources provide a rich repository of accurate and up-to-date information, ensuring a broad and reliable foundation for incorporating external biomedical knowledge.
    \item \textbf{Hierarchical Indexing Strategy.}
    For the Bio-RE task, we propose a Hierarchical Indexing Strategy. Traditional indexing strategies often rely on simple chunking methods \cite{efeoglu2024retrieval_augmented,denseRetrieval,2024-Retrieval-AugmentedBlack-Box_Language_Models}. However, due to the synonymy and aliases of biomedical entities, as well as the vastness of biomedical databases, these chunking strategies no longer meet the requirements of the Bio-RE task. Inspired by the CUIs from the UMLS, we propose an indexing strategy that combines CUIs with chunking, ensuring more precise and comprehensive indexing for biomedical data. 
    Specifically, we construct a hierarchical indexing structure by first indexing the CUIs of biomedical entities as the primary layer. Next, we assign each document related to an entity to its corresponding CUI index and then further index the document chunks as the secondary layer.
    
    Using CUIs instead of entity names as indexes mitigates the effects of synonymy and aliases in biomedical entities. CUIs serve as unique identifiers that consolidate synonyms and alternative terms for the same concept, reducing inconsistencies arising from varied terminologies. This approach thus enhances retrieval accuracy and relevance, particularly in complex biomedical contexts where entities often have aliases or ambiguous meanings.
    \item \textbf{CUI Retrieval and Generation.}
\begin{figure}[!t]
\centering
\includegraphics[width=0.9\columnwidth]{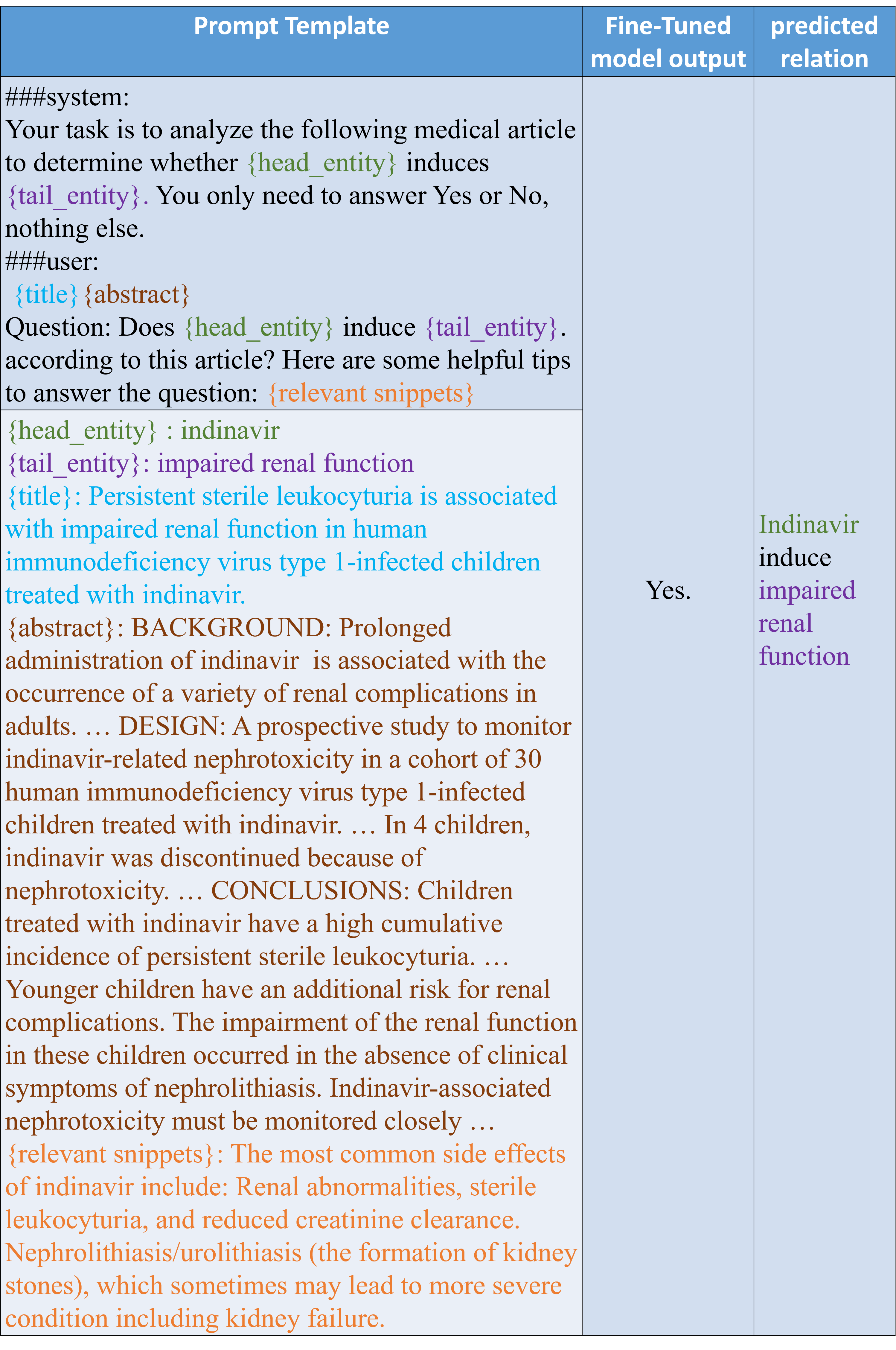}
\caption{An example of the final inference prompt from the CDR dataset. The fine-tuned model takes this prompt as input and outputs the predicted relation.}
\label{fig_4}
\end{figure}
    We use an embedding model to convert the document chunks into vectors, creating a hierarchical vector structure with a similar organization. The retriever, using the input head and tail biomedical entities $(h_{i,j}, t_{i,j})$ along with their corresponding CUIs, searches this hierarchical vector structure to locate the relevant document chunk vectors. 
    Next, it selects the relevant biomedical snippets $dr_{i,j}$ based on cosine similarity.
\begin{table}[!t] 
	       
        \caption{Statistics of the datasets.} 
	\begin{center}                 
		\begin{tabular}{c c c c c} 
			\toprule              
			Statistics / Dataset &CDR&GDA&BioRED\\       
			\hline                 
			\# Train &500&23353&400 \\
			\# Dev &500&5839&100\\
                \# Test &500&1000&100\\
                \# Relation types &2&2&8\\
                Avg.\# relations per Doc. &2.1&1.6&10.8\\
			\bottomrule            
		\end{tabular}

        \label{table1}
	\end{center}
\end{table}
    These relevant biomedical snippets $dr_{i,j}$ are combined with the original input to form the final inference prompt, which is fed into the fine-tuned model $\widetilde{M}$  to obtain the predicted relation $\hat{r}_{i,j}$. This process can be formally described as follows:
    \begin{equation}
    \begin{split}
    \hat{r}_{i,j}=\widetilde{M}(I,d_i,dr_{i,j},h_{i,j}, t_{i,j})
    \end{split}
    \end{equation}
    In this equation, $i$ denotes the index of the sample in the dataset, and $j$ represents the $j$-th entity pair within that sample. $\hat{r}_{i,j}$ represents the predicted relation.
    The fine-tuned model $\widetilde{M}$ receives the task instruction $I$, the original document $d_i$, the relevant biomedical snippets $dr_{i,j}$ and the head and tail entities $(h_{i,j}, t_{i,j})$ as inputs. An example of this process is illustrated in the Figure \ref{fig_4}.

    Compared to traditional RAG methods, CUI RAG leverages CUIs to restrict the retrieval scope to documents specifically focused on the head and tail entities, significantly narrowing the search range, improving retrieval relevance, and reducing the impact of entity synonymy and aliases on retrieval.

\end{itemize}

\section{EXPERIMENTS}

\subsection{Datasets}
We evaluated our framework on three public document-level Bio-RE datasets: CDR, GDA, and BioRED. The dataset statistics are shown in Table \ref{table1}.

\textbf{CDR} \cite{10.1093/database/CDR}.
The Chemical-Disease Reactions (CDR) dataset, constructed from PubMed abstracts, contains 1,500 human-annotated documents divided equally into training, development, and test sets. It focuses on the binary classification task of identifying Chemical-Induced-Disease relations between chemical and disease entities.

\textbf{GDA} \cite{10.1007/GDA}.
The Gene-Disease Associations (GDA) dataset is a large-scale biomedical dataset constructed from MEDLINE abstracts using distant supervision. Following Christopoulou et al. \cite{christopoulou-etal-2019-connecting}, we split the training set into 23,353 training documents and 5,839 development documents. The primary task is to predict binary interactions between Gene and Disease entities.

\textbf{BioRED} \cite{BioRED}.
Unlike previous datasets that focus only on binary relations and a single entity pair, the biomedical relation extraction dataset (BioRED) includes various entity types such as gene, disease, chemical, variant, species, and cell line. It also encompasses multiple relation pairs (e.g., gene-disease, chemical-chemical) and various types of relations.

\begin{table*}[t]
\centering
\caption{Results on the test set of CDR and GDA. We categorized the baseline models into three groups: graph-based models, transformer-based models, and LLM-based models.}
\label{table2}
\begin{tabular}{cccccccc}
\toprule
 &\multicolumn{3}{c}{CDR $F_{1}$(\%)}  & &\multicolumn{3}{c}{GDA $F_{1}$(\%)}    \\
\cmidrule(lr){2-4} \cmidrule(lr){6-8}
&Overall&Intra-&Inter-&  &Overall&Intra-&Inter-\\
\midrule
\multicolumn{8}{c}{\textbf{Graph-based model}}\\
\specialrule{.1em}{.2em}{.2em}
CGM2IR\cite{Inter_Pair_Reasoning2023}&73.8 &79.2 &55.1 & &84.7 &88.3 &59.0\\
FILR\cite{li-etal-2022-document}&85.7&89.1&77.2&  &84.7&87.2 &68.9\\
HTGRS\cite{HTGRS}& 86.9&90.9&75.1&  &87.3&89.2 &69.7\\
FCDS\cite{zhu2024fcdsfusingconstituencydependency}&72.6&-&-&  & 87.4&-&-\\
Topic-BiGRU-U-Net\cite{Topic-BiGRU-U-Net}&87.1&89.4&81.7&  &  84.1&86.7&68.3\\
\specialrule{.1em}{.2em}{.2em}
\multicolumn{8}{c}{\textbf{Transformer-based model}} \\
\specialrule{.1em}{.2em}{.2em}
TriA-BioRE\cite{TriA-BioRE}&65.0 &- &- & &83.8 &- &-\\
SSAN\cite{xu2021Entitystructurewithinandthroughout}&68.7 &74.5 &56.2 & &83.7&86.6 &65.3\\
ALOTP\cite{zhou2021documentALOTP}&69.4 &- &- & &83.9&- &-\\
DocRE-II\cite{zhang-etal-2022-IterativeInference}&73.2 &- &- & &85.9 &- &-\\
DocRE-SD\cite{zhang2023self-distillation}&76.8 &- &- & &86.4 &- &-\\
SAIS\cite{xiao-etal-2022-sais}&79.0 &- &- & &87.1 &- &-\\

PSD\cite{DocRE_PSD}&86.1 &89.3 &78.7 & &84.9 &87.4 &66.7\\

\specialrule{.1em}{.2em}{.2em}
\multicolumn{8}{c}{\textbf{LLM-based model}}\\
\specialrule{.1em}{.2em}{.2em}
Multi-Span\cite{LLMDOCRE}&71.2 &75.3 &56.7 & &85.2 &88.6 &62.7\\
LLaMA2-7B-Chat&72.1 &77.1 &63.0 & &69.4 &76.7&44.8\\
GPT-3.5&70.8 &74.7 &62.4 & &60.9 &67.0&36.6\\
GPT-4&80.0 &85.2 &72.5 & &64.6 &70.9 &39.7\\
\textbf{Ours}&\textbf{88.2} &\textbf{90.8} &\textbf{82.3} & &\textbf{88.7} &\textbf{90.9} &\textbf{77.1}\\

\bottomrule
\end{tabular}
\end{table*}
\subsection{Experimental Settings}

During the ADRCM fine-tuning stage, we set the threshold $\beta$ for IoRs to 3 and utilized the GPT-3.5-turbo-0125 API for ChatGPT. LLaMA2-7B-Chat was selected as the backbone model, and the PEFT method, LoRA, was employed. 
For the CDR dataset, we set the LoRA decomposition rank to 16 and LoRA alpha to 32. For the GDA and BioRED datasets, we set the LoRA decomposition rank to 64 and LoRA alpha to 16.
Across all three datasets, a learning rate of 2e-4 and a LoRA dropout rate of 0.1 were used. During the inference stage, jina-embeddings-v2-base-en was chosen as the embedding model \cite{günther2023jinaEmbeddings}.
This model uses Attention with Linear Biases instead of traditional positional embeddings to efficiently encode extended text sequences while maintaining strong performance. Additionally, it supports a sequence length of up to 8192 tokens.

\subsection{Experimental Results}

\subsubsection{CDR and GDA Results}
We conducted comprehensive and comparative experiments on the CDR and GDA datasets, with the results presented in Table \ref{table2}. The baseline models are categorized into three groups: graph-based, transformer-based, and LLM-based models.

Graph-based models include CGM2IR \cite{Inter_Pair_Reasoning2023}, FILR \cite{li-etal-2022-document},
HTGRS \cite{HTGRS}, FCDS \cite{zhu2024fcdsfusingconstituencydependency}, and Topic-BiGRU-U-Net \cite{Topic-BiGRU-U-Net}. 
Transformer-based models include TriA-BioRE\cite{TriA-BioRE}, SSAN \cite{xu2021Entitystructurewithinandthroughout}, ALOTP \cite{zhou2021documentALOTP}, DocRE-II \cite{zhang-etal-2022-IterativeInference}, DocRE-SD \cite{zhang2023self-distillation}, SAIS \cite{xiao-etal-2022-sais}, and PSD \cite{DocRE_PSD}.
LLM-based models include Multi-Span \cite{LLMDOCRE}, LLaMA2-7B-Chat, GPT-3.5, and GPT-4.

As shown in Table \ref{table2}, our framework (Ours) demonstrates significant improvements in the CDR and GDA datasets, achieving new state-of-the-art performance.

On the CDR dataset, our framework achieves  overall $F_{1}$ of 88.2\%,  Intra-$F_{1}$ of 90.8\%, and  Inter-$F_{1}$ of 82.3\%. This performance surpasses the current state-of-the-art graph-based model, Topic-BiGRU-U-Net, by 1.1\% in overall $F_{1}$. Compared to the powerful GPT-4 model, our framework shows an improvement of 8.2\% in overall $F_{1}$, and compared to the backbone model LLaMA2-7B-Chat, it achieves a substantial enhancement of 16.1\%.

On the GDA dataset, 
our framework attains overall $F_{1}$ of 88.7\%,  Intra-$F_{1}$ of 90.9\%, and Inter-$F_{1}$ of 77.1\%.
This represents a 1.3\% improvement over FCDS, 24.1\% improvement over GPT-4, and 19.3\% improvement compared to the backbone model LLaMA2-7B-Chat. 
Additionally, we observe that our framework exhibits a notable enhancement in Inter-$F_{1}$.
On the CDR dataset, it outperforms the backbone model LLaMA2-7B-Chat by 19.3\% and GPT-4 by 9.8\%.
Similarly, on the GDA dataset, our framework demonstrates a remarkable improvement, surpassing LLaMA2-7B-Chat by 32.3\% and GPT-4 by an even more substantial 37.4\%, and outperforming HTGRS by 7.4\%. Notably, it also achieves approximately 10\% improvement over most graph-based and transformer-based models.
The significant performance improvement is primarily attributed to the effectiveness of ADRCM fine-tuning. ADRCM enables the model to focus on the most relevant information for each specific relation, while also allowing it to capture and distinguish critical relational cues across sentences.
The observed gains on the CDR and GDA datasets underscore that ADRCM fine-tuning strengthens the model's cross-sentence inference capabilities, enabling it to better understand complex biomedical relations and achieve superior performance compared to other models.
\subsubsection{BioRED Results}
\begin{table}[!t]
    \caption{Results on the test set of BioRED.}
    \centering
    \begin{tabular}{cccc}
    \toprule
         Model& Precision(\%)& Recall(\%)& $F_{1}$(\%)\\
    \midrule
         TriA-BioRE\cite{TriA-BioRE}& 61.7 &42.4 &50.3\\
         BERT-GT\cite{BertGT}&55.0&58.7&56.8 \\  
         PubMedBERT\cite{pubmedBert}&54.2&63.8&58.6 \\
         ATLOP\cite{zhou2021documentALOTP}&58.7&68.4 &63.1 \\
         SAIS\cite{xiao-etal-2022-sais}&60.5&67.1 &63.8 \\
         HTGRS\cite{HTGRS}&59.3&\textbf{76.8} &66.9 \\
     \specialrule{.1em}{.2em}{.2em}
\multicolumn{4}{c}{\textbf{LLM-based model}}\\
\specialrule{.1em}{.2em}{.2em}   
         LLaMA2-7B-Chat&  21.5 &29.7 &24.9  \\
         GPT-3.5&  47.3 &39.1 &42.8  \\
         GPT-4&  39.8 &56.7 &46.8  \\
         \textbf{Ours}& \textbf{81.5} &65.6&\textbf{72.7}  \\
    \bottomrule
    \end{tabular}
   
    \label{table3}
\end{table}
The CDR and GDA datasets have relatively limited types of relations and entities. To further evaluate the performance of our framework in scenarios involving multiple entity types, multiple relation types, and a higher density of information (with more relations per document on average), we conducted experiments on the BioRED dataset. 
The results of these experiments are presented in Table \ref{table3}.
We compare the performance of our framework against nine baseline models: TriA-BioRE \cite{TriA-BioRE}, BERT-GT \cite{BertGT}, PubMedBERT \cite{pubmedBert}, ATLOP \cite{zhou2021documentALOTP}, SAIS \cite{xiao-etal-2022-sais}, HTGRS \cite{HTGRS}, LLaMA2-7B-Chat, GPT-3.5, and GPT-4.
As shown in Table \ref{table3}, our framework achieves $F_{1}$ of 72.7\%, demonstrating state-of-the-art performance on the BioRED dataset, consistent with results on the previous two datasets. 
Although HTGRS has the highest recall, its $F_{1}$ is lower due to a relatively low precision. In contrast, our framework sets a new benchmark by surpassing HTGRS by 5.8\% and outperforming GPT-4 by 25.9\% in $F_{1}$. Additionally, compared to the backbone model LLaMA2-7B-Chat, our framework achieves a substantial improvement, increasing  $F_{1}$ from 24.9\% to 72.7\%.
These results underscore the exceptional performance and robustness of our framework in handling information-dense datasets with diverse relations.

\subsection{Effectiveness of IoRs }
\begin{figure}[!t]
\centering
\includegraphics[width=0.9\columnwidth]{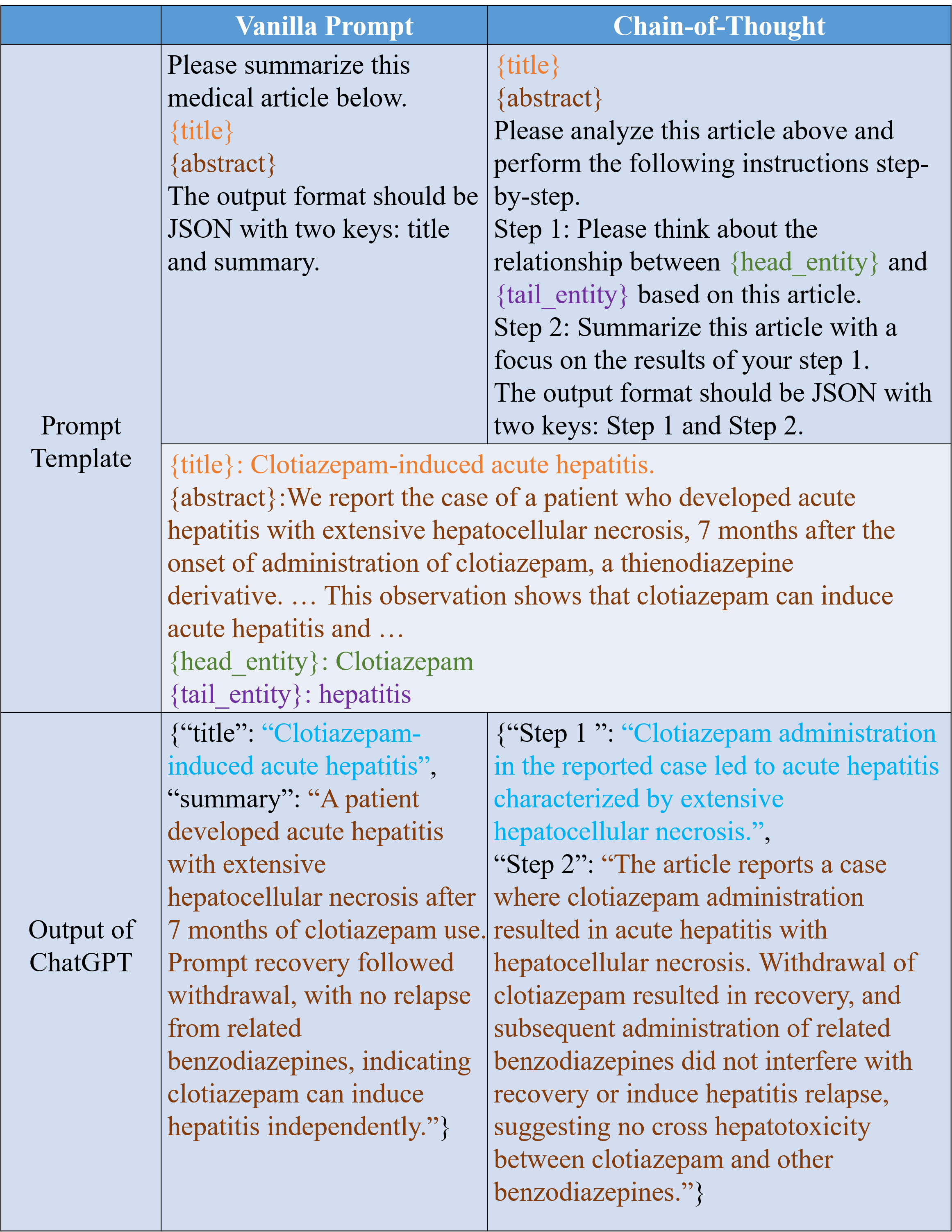}
\caption{An example of a vanilla and chain-of-thought prompt. Our proposed IoRs prompt is illustrated in Figure \ref{fig_3}. Using these three types of prompts, we generated three distinct sets of synthetic data with ChatGPT.}
\label{fig_5}
\end{figure}
To further assess the impact of our proposed IoRs prompt, we generated synthetic data using three different prompts: IoRs prompt, vanilla prompt, and chain-of-thought prompt, as illustrated in Figure \ref{fig_5}. To ensure the fairness of the experiment, we randomly sampled 487 examples from each dataset produced by these prompts. The LLaMA2-7B-Chat model was then fine-tuned using the synthetic data generated by each prompt, and its performance was evaluated on the CDR and GDA datasets.

\begin{table*}[!t]
\caption{Experimental results on the CDR and GDA datasets using LLaMA2-7B-Chat fine-tuned with data generated by the Vanilla Prompt, Chain-of-Thought, and Iteration-of-REsummary.}
\centering

\begin{tabular}{cccccccc}
\toprule
 &\multicolumn{3}{c}{CDR $F_{1}$(\%)}   & &\multicolumn{3}{c}{GDA $F_{1}$(\%)}    \\
\cmidrule(lr){2-4} \cmidrule(lr){6-8}
&Overall&Intra-&Inter- & &Overall&Intra-&Inter-\\
\midrule
Vanilla Prompt&77.6 &82.1 &66.9 & &73.3 &77.2 &48.6\\
Chain-of-Thought&78.6 &82.3 &69.7&  &76.4 &80.1 &54.4\\
\textbf{Iteration-of-REsummary}
&\textbf{80.0}&\textbf{84.1} &\textbf{70.4} & &\textbf{80.7} &\textbf{84.5} &\textbf{61.2}\\
\bottomrule
\end{tabular}

\label{table4}
\end{table*}
As shown in Table \ref{table4}, the LLaMA2-7B-Chat model, fine-tuned using synthetic data generated by the IoRs prompt, achieves overall $F_{1}$ of 80.0\%, Intra-$F_{1}$ of 84.1\%, and Inter-$F_{1}$ of 70.4\% on the CDR dataset. On the GDA dataset, it achieves overall $F_{1}$  of 80.7\%, Intra-$F_{1}$ of 84.5\%, and Inter-$F_{1}$ of 61.2\%. 
Furthermore, it outperforms the model fine-tuned with data from the chain-of-thought prompt by 1.4\% on the CDR dataset and 4.3\% on the GDA dataset in $F_{1}$.  It also surpasses the model fine-tuned with data from the vanilla prompt by 2.4\% on the CDR dataset and 7.4\% on the GDA dataset in $F_{1}$. Additionally, it achieves superior performance in both Intra-$F_{1}$ and Inter-$F_{1}$. 
Based on our analysis and observations, IoRs outperforms Chain of Thought for two key reasons. First, Chain of Thought suffers from error propagation, caused by an incorrect relation identified in the initial step. Second, its summaries in the second step sometimes fail to focus on the specific entity pair and their relation. As shown in Figure \ref{fig_5}, Chain of Thought focuses more on the relation between \textit{clotiazepam} and \textit{benzodiazepines} rather than the intended head and tail entities, \textit{clotiazepam} and \textit{hepatitis}. In contrast, our proposed IoRs effectively address these issues. Through relation confirmation, it ensures that the generated summary corresponds to the original relation, and by iteratively refining mismatched summaries, it enables the model to concentrate on the specific entity pair and their relation.
\subsection{Ablation Study}
\begin{table*}[!t]
\caption{Ablation study of our framework on the test set of CDR and GDA, where P represents Precision and R represents Recall.}
\centering

\begin{tabular}{lcccccccccc}
\toprule

&\multicolumn{5}{c}{CDR metrics(\%)}& \multicolumn{5}{c}{GDA metrics(\%)}\\
\cmidrule(lr){2-6} \cmidrule(lr){7-11}
&Overall $F_{1}$&Intra-$F_{1}$&Inter-$F_{1}$&P&R
&Overall $F_{1}$&Intra-$F_{1}$&Inter-$F_{1}$&P&R\\
\midrule
w/o synthetic data &85.3 &88.6 &77.7&79.0   &92.7 &86.3 &88.8 &74.3&82.5 &90.5 \\
w/o ADRCM fine-tuning&77.0&81.8&67.6 &65.3  &94.0 &73.2&80.3 &48.1&60.0 &93.9\\
fine-tuning w/o ADRCM&69.9&76.9&57.9&53.9 &\textbf{99.7} &71.6&79.1&47.8&55.9 &\textbf{99.4}\\
w/o CUI RAG&84.9&87.8&78.1 &82.5  &87.3 & 87.7&89.7&76.9&82.9 &93.1\\
RAG w/o CUI &82.4&85.5&75.9&76.5 &89.2  &85.5&88.4&69.4&84.6 &86.4\\
LLaMA2-7B-Chat &72.1&77.1&63.0&60.0 &90.4 &69.4&76.7&44.8&56.4 &90.2\\
\textbf{Ours}&\textbf{88.2} &\textbf{90.8} &\textbf{82.3} &\textbf{83.4}  &93.6 &\textbf{88.7} &\textbf{90.9} &\textbf{77.1}&\textbf{83.6} &94.3\\
\bottomrule
\end{tabular}

\label{table5}
\end{table*}
To analyze the role and impact of each component of our framework, we conducted an ablation study focusing on three key components: synthetic data, ADRCM fine-tuning, and CUI RAG.

As shown in Table \ref{table5}, the performance decreases with the removal of each component, demonstrating the contribution and importance of every element in our framework. 
Specifically, the removal of synthetic data during fine-tuning results in 2.9\% $F_{1}$ decrease on the CDR dataset and 2.4\% $F_{1}$ decrease on the GDA dataset. This highlights the significant impact of synthetic data generated by the IoRs prompt. 
When we skip ADRCM fine-tuning and use the backbone model with CUI RAG for inference, we observe a substantial performance drop of 11.2\% $F_{1}$ on the CDR dataset and 15.5\% $F_{1}$ on the GDA dataset, with an even more pronounced decline in Inter-$F_{1}$ of 14.7\% and 29\%, respectively. This further demonstrates the critical role of ADRCM fine-tuning in improving the model's cross-sentence inference capabilities.

To further validate the impact of ADRCM, we conducted an experiment in which ADRCM was removed during fine-tuning, using only the original training data and synthetic data. In this scenario, the model predominantly predicts positive relations, leading to a recall close to 100\%. This outcome highlights the critical role of ADRCM in the fine-tuning process, indicating that the improvements achieved with ADRCM fine-tuning are specifically due to ADRCM itself, rather than the fine-tuning process.

Furthermore, directly using the ADRCM fine-tuned model for inference without CUI RAG results in a 3.3\% $F_{1}$ decrease on the CDR dataset and a 1\% $F_{1}$ decrease on the GDA dataset.
Combined with the comparisons in the second (w/o ADRCM fine-tuning) and sixth rows (LLaMA2-7B-Chat), CUI RAG enhances performance by increasing $F_{1}$ by 4.9\% on the CDR dataset and 3.8\% on the GDA dataset. These results suggest that our CUI RAG enhances retrieval relevance and supplies the model with valuable information, thereby aiding in solving the Bio-RE task.
Finally, the removal of CUI in RAG, with only the chunking strategy used during inference, leads to 5.8\% decrease in $F_{1}$ on the CDR dataset and 3.2\% decrease in the GDA dataset. 
Notably, $F_{1}$ of RAG without CUI is even lower than that of without CUI RAG.
Based on our observations, the cause of this outcome is the inherent polysemy and aliases of biomedical entities. The chunking strategy, which relies solely on text matching, often retrieves documents containing a significant amount of information unrelated to the head and tail entities being predicted. This negatively impacts the model's performance by introducing irrelevant information. In contrast, CUI RAG, by incorporating CUIs and the Hierarchical Indexing Strategy, mitigates the effects of entity polysemy and aliases in retrieval and narrows the search scope to documents specifically centered on the head and tail entities, effectively avoiding these issues.

\section{Conclusion}

In this paper, we propose a novel framework for document-level Bio-RE via LLM Adaptive Document-Relation Cross-Mapping  fine-tuning and Concept Unique Identifier RAG. Experimental results on the CDR, GDA, and BioRED datasets demonstrate that our framework achieves state-of-the-art performance across all three datasets.
However, our framework requires initializing a predefined set of relation types and faces challenges when dealing with a large number of relation types. Moreover, in the CUI RAG, we narrow the retrieval scope to documents focused on the head and tail entities, which may lead to some useful information being overlooked. 
In future work, we aim to enable Bio-RE without relying on a predefined set of relation types, thereby improving the framework's ability to effectively handle scenarios with numerous relation types. Additionally, we plan to improve the retrieval strategy in CUI RAG by dynamically expanding the scope beyond documents focused on the head and tail entities, which will allow for broader contextual information and reduce the likelihood of overlooking valuable content.

\bibliographystyle{IEEEtran}
\bibliography{cite}

\begin{thebibliography}{10}
\providecommand{\url}[1]{#1}
\csname url@samestyle\endcsname
\providecommand{\newblock}{\relax}
\providecommand{\bibinfo}[2]{#2}
\providecommand{\BIBentrySTDinterwordspacing}{\spaceskip=0pt\relax}
\providecommand{\BIBentryALTinterwordstretchfactor}{4}
\providecommand{\BIBentryALTinterwordspacing}{\spaceskip=\fontdimen2\font plus
\BIBentryALTinterwordstretchfactor\fontdimen3\font minus \fontdimen4\font\relax}
\providecommand{\BIBforeignlanguage}[2]{{%
\expandafter\ifx\csname l@#1\endcsname\relax
\typeout{** WARNING: IEEEtran.bst: No hyphenation pattern has been}%
\typeout{** loaded for the language `#1'. Using the pattern for}%
\typeout{** the default language instead.}%
\else
\language=\csname l@#1\endcsname
\fi
#2}}
\providecommand{\BIBdecl}{\relax}
\BIBdecl

\bibitem{Distantly-Supervised-RE}
T.~Liang, Y.~Liu, X.~Liu, H.~Zhang, G.~Sharma, and M.~Guo, ``Distantly-supervised long-tailed relation extraction using constraint graphs,'' \emph{IEEE Transactions on Knowledge and Data Engineering}, vol.~35, no.~7, pp. 6852--6865, 2023.

\bibitem{FPrompt-PLM}
L.~Zhang, Y.~Li, Q.~Wang, Y.~Wang, H.~Yan, J.~Wang, and J.~Liu, ``Fprompt-plm: Flexible-prompt on pretrained language model for continual few-shot relation extraction,'' \emph{IEEE Transactions on Knowledge and Data Engineering}, pp. 1--15, 2024.

\bibitem{KnowPrompt}
X.~Chen, N.~Zhang, X.~Xie, S.~Deng, Y.~Yao, C.~Tan, F.~Huang, L.~Si, and H.~Chen, ``Knowprompt: Knowledge-aware prompt-tuning with synergistic optimization for relation extraction,'' in \emph{Proceedings of the ACM Web Conference 2022}, ser. WWW '22, 2022, pp. 2778--2788.

\bibitem{Biomedical_Distant_SupervisionRE}
H.~Zhang, Y.~Liu, X.~Liu, T.~Liang, G.~Sharma, L.~Xue, and M.~Guo, ``Sentence bag graph formulation for biomedical distant supervision relation extraction,'' \emph{IEEE Transactions on Knowledge and Data Engineering}, vol.~36, no.~9, pp. 4890--4903, 2024.

\bibitem{REBEL}
P.-L. Huguet~Cabot and R.~Navigli, ``{REBEL}: Relation extraction by end-to-end language generation,'' in \emph{Findings of the Association for Computational Linguistics: EMNLP 2021}, 2021, pp. 2370--2381.

\bibitem{LearningRelationPrototype}
Y.~Cao, J.~Kuang, M.~Gao, A.~Zhou, Y.~Wen, and T.-S. Chua, ``Learning relation prototype from unlabeled texts for long-tail relation extraction,'' \emph{IEEE Transactions on Knowledge and Data Engineering}, vol.~35, no.~2, pp. 1761--1774, 2023.

\bibitem{yao-etal-2019-docred}
Y.~Yao, D.~Ye, P.~Li, X.~Han, Y.~Lin, Z.~Liu, Z.~Liu, L.~Huang, J.~Zhou, and M.~Sun, ``{D}oc{RED}: A large-scale document-level relation extraction dataset,'' in \emph{Proceedings of the 57th Annual Meeting of the Association for Computational Linguistics}, 2019, pp. 764--777.

\bibitem{TKDEDocRE}
T.~Xu, J.~Qu, W.~Hua, Z.~Li, J.~Xu, A.~Liu, L.~Zhao, and X.~Zhou, ``Evidence reasoning and curriculum learning for document-level relation extraction,'' \emph{IEEE Transactions on Knowledge and Data Engineering}, vol.~36, no.~2, pp. 594--607, 2024.

\bibitem{10.1007/GDA}
Y.~Wu, R.~Luo, H.~C.~M. Leung, H.-F. Ting, and T.~W. Lam, ``Renet: A deep learning approach for extracting gene-disease associations from literature,'' in \emph{Annual International Conference on Research in Computational Molecular Biology}, 2019, pp. 272--284.

\bibitem{Wikidata}
D.~Vrande{\v{c}}i{\'c} and M.~Kr{\"o}tzsch, ``Wikidata: a free collaborative knowledgebase,'' \emph{Communications of the ACM}, vol.~57, no.~10, pp. 78--85, 2014.

\bibitem{10.1093/database/CDR}
J.~Li, Y.~Sun, R.~J. Johnson, D.~Sciaky, C.-H. Wei, R.~Leaman, A.~P. Davis, C.~J. Mattingly, T.~C. Wiegers, and Z.~Lu, ``Biocreative v cdr task corpus: a resource for chemical disease relation extraction,'' \emph{Database: The Journal of Biological Databases and Curation}, vol. 2016, p. baw068, 2016.

\bibitem{openai2024gpt4}
J.~Achiam, S.~Adler, S.~Agarwal, L.~Ahmad, I.~Akkaya, F.~L. Aleman, D.~Almeida, J.~Altenschmidt, S.~Altman, S.~Anadkat \emph{et~al.}, ``Gpt-4 technical report,'' \emph{arXiv preprint arXiv:2303.08774}, 2023.

\bibitem{LLaMA2}
H.~Touvron, L.~Martin, K.~Stone, P.~Albert, A.~Almahairi, Y.~Babaei, N.~Bashlykov, S.~Batra, P.~Bhargava, S.~Bhosale \emph{et~al.}, ``Llama 2: Open foundation and fine-tuned chat models,'' \emph{arXiv preprint arXiv:2307.09288}, 2023.

\bibitem{wei2024chatie}
X.~Wei, X.~Cui, N.~Cheng, X.~Wang, X.~Zhang, S.~Huang, P.~Xie, J.~Xu, Y.~Chen, M.~Zhang \emph{et~al.}, ``Chatie: Zero-shot information extraction via chatting with chatgpt,'' \emph{arXiv preprint arXiv:2302.10205}, 2023.

\bibitem{wan2023gpt}
Z.~Wan, F.~Cheng, Z.~Mao, Q.~Liu, H.~Song, J.~Li, and S.~Kurohashi, ``Gpt-re: In-context learning for relation extraction using large language models,'' in \emph{Proceedings of the 2023 Conference on Empirical Methods in Natural Language Processing (EMNLP)}, 2023, pp. 3534--3547.

\bibitem{Xue_AutoRE}
X.~Lilong, Z.~Dan, D.~Yuxiao, and T.~Jie, ``Autore: Document-level relation extraction with large language models,'' \emph{arXiv preprint arXiv:2403.14888}, 2024.

\bibitem{BioRED}
L.~Luo, P.-T. Lai, C.-H. Wei, C.~N. Arighi, and Z.~Lu, ``Biored: a rich biomedical relation extraction dataset,'' \emph{Briefings in Bioinformatics}, vol.~23, no.~5, p. bbac282, 2022.

\bibitem{christopoulou-etal-2019-connecting}
F.~Christopoulou, M.~Miwa, and S.~Ananiadou, ``Connecting the dots: Document-level neural relation extraction with edge-oriented graphs,'' in \emph{Proceedings of the 2019 Conference on Empirical Methods in Natural Language Processing and the 9th International Joint Conference on Natural Language Processing (EMNLP-IJCNLP)}, 2019, pp. 4925--4936.

\bibitem{Nan_Guo_Sekulic_Lu_2020}
G.~Nan, Z.~Guo, I.~Sekulic, and W.~Lu, ``Reasoning with latent structure refinement for document-level relation extraction,'' in \emph{Proceedings of the 58th Annual Meeting of the Association for Computational Linguistics}, 2020, pp. 1546--1557.

\bibitem{peng2022document}
X.~Peng, C.~Zhang, and K.~Xu, ``Document-level relation extraction via subgraph reasoning,'' in \emph{Proceedings of the Thirty-First International Joint Conference on Artificial Intelligence, {IJCAI-22}}, 2022, pp. 4331--4337.

\bibitem{Inter_Pair_Reasoning2023}
D.~Zeng, C.~Zhao, C.~Jiang, J.~Zhu, and J.~Dai, ``Document-level relation extraction with context guided mention integration and inter-pair reasoning,'' \emph{IEEE/ACM Transactions on Audio, Speech, and Language Processing}, vol.~31, pp. 3659--3666, 2023.

\bibitem{Document_Level_Relation_Extraction_Path_Reasoning2023}
W.~Xu, K.~Chen, and T.~Zhao, ``Document-level relation extraction with path reasoning,'' \emph{ACM Transactions on Asian and Low-Resource Language Information Processing}, vol.~22, no.~4, pp. 1--14, 2023.

\bibitem{Reconstruction2021}
------, ``Document-level relation extraction with reconstruction,'' in \emph{Proceedings of the AAAI Conference on Artificial Intelligence}, vol.~35, no.~16, 2021, pp. 14\,167--14\,175.

\bibitem{DAGCN}
D.~Zhang, Z.~Liu, W.~Jia, F.~Wu, H.~Liu, and J.~Tan, ``Dual attention graph convolutional network for relation extraction,'' \emph{IEEE Transactions on Knowledge and Data Engineering}, vol.~36, no.~2, pp. 530--543, 2024.

\bibitem{tian-etal-2021-dependency}
Y.~Tian, G.~Chen, Y.~Song, and X.~Wan, ``Dependency-driven relation extraction with attentive graph convolutional networks,'' in \emph{Proceedings of the 59th Annual Meeting of the Association for Computational Linguistics and the 11th International Joint Conference on Natural Language Processing (Volume 1: Long Papers)}, 2021, pp. 4458--4471.

\bibitem{hang2024graphAtt}
T.~Hang, J.~Feng, Y.~Wang, and L.~Yan, ``Graph neural networks with selective attention and path reasoning for document-level relation extraction,'' \emph{Applied Intelligence}, pp. 1--20, 2024.

\bibitem{Topic-BiGRU-U-Net}
Y.~Zhao and R.~Yan, ``Topic-bigru-u-net for document-level relation extraction from biomedical literature,'' in \emph{2023 IEEE International Conference on Bioinformatics and Biomedicine (BIBM)}, 2023, pp. 1000--1003.

\bibitem{li-etal-2022-document}
L.~Li, R.~Lian, H.~Lu, and J.~Tang, ``Document-level biomedical relation extraction based on multi-dimensional fusion information and multi-granularity logical reasoning,'' in \emph{Proceedings of the 29th International Conference on Computational Linguistics}, 2022, pp. 2098--2107.

\bibitem{AGCN}
C.~Park, J.~Park, and S.~Park, ``Agcn: Attention-based graph convolutional networks for drug-drug interaction extraction,'' \emph{Expert Systems with Applications}, vol. 159, p. 113538, 2020.

\bibitem{Dual-Attention_bioRE}
L.~Li, R.~Lian, and H.~Lu, ``Document-level biomedical relation extraction with generative adversarial network and dual-attention multi-instance learning,'' in \emph{2021 IEEE International Conference on Bioinformatics and Biomedicine (BIBM)}, 2021, pp. 438--443.

\bibitem{HTGRS}
J.~Yuan, F.~Zhang, Y.~Qiu, H.~Lin, and Y.~Zhang, ``{Document-level biomedical relation extraction via hierarchical tree graph and relation segmentation module},'' \emph{Bioinformatics}, vol.~40, no.~7, p. btae418, 2024.

\bibitem{xu2021Entitystructurewithinandthroughout}
B.~Xu, Q.~Wang, Y.~Lyu, Y.~Zhu, and Z.~Mao, ``Entity structure within and throughout: Modeling mention dependencies for document-level relation extraction,'' \emph{Proceedings of the AAAI Conference on Artificial Intelligence}, vol.~35, no.~16, pp. 14\,149--14\,157, 2021.

\bibitem{zhou2021documentALOTP}
W.~Zhou, K.~Huang, T.~Ma, and J.~Huang, ``Document-level relation extraction with adaptive thresholding and localized context pooling,'' \emph{Proceedings of the AAAI Conference on Artificial Intelligence}, vol.~35, no.~16, pp. 14\,612--14\,620, 2021.

\bibitem{zhang-etal-2022-IterativeInference}
L.~Zhang, J.~Su, Y.~Chen, Z.~Miao, M.~Zijun, Q.~Hu, and X.~Shi, ``Towards better document-level relation extraction via iterative inference,'' in \emph{Proceedings of the 2022 Conference on Empirical Methods in Natural Language Processing (EMNLP)}, 2022, pp. 8306--8317.

\bibitem{xiao-etal-2022-sais}
Y.~Xiao, Z.~Zhang, Y.~Mao, C.~Yang, and J.~Han, ``{SAIS}: Supervising and augmenting intermediate steps for document-level relation extraction,'' in \emph{Proceedings of the 2022 Conference of the North American Chapter of the Association for Computational Linguistics: Human Language Technologies}, 2022, pp. 2395--2409.

\bibitem{zhang2023self-distillation}
L.~Zhang, J.~Su, Z.~Min, Z.~Miao, Q.~Hu, B.~Fu, X.~Shi, and Y.~Chen, ``Exploring self-distillation based relational reasoning training for document-level relation extraction,'' \emph{Proceedings of the AAAI Conference on Artificial Intelligence}, vol.~37, no.~11, pp. 13\,967--13\,975, 2023.

\bibitem{TriA-BioRE}
L.~Chen, J.~Su, T.-W. Lam, and R.~Luo, ``Exploring pair-aware triangular attention for biomedical relation extraction,'' in \emph{Proceedings of the 14th ACM International Conference on Bioinformatics, Computational Biology, and Health Informatics}, 2023, pp. 1--5.

\bibitem{LLMDOCRE}
X.~Wang, J.~Liu, J.~Wang, J.~Duan, G.~Guan, Q.~Zhang, and J.~Zhou, ``Document-level relation extraction based on machine reading comprehension and hybrid pointer-sequence labeling,'' \emph{ACM Transactions on Asian and Low-Resource Language Information Processing}, vol.~23, no.~7, pp. 1--16, 2024.

\bibitem{guo2023code-generationUIE}
Y.~Guo, Z.~Li, X.~Jin, Y.~Liu, Y.~Zeng, W.~Liu, X.~Li, P.~Yang, L.~Bai, J.~Guo \emph{et~al.}, ``Retrieval-augmented code generation for universal information extraction,'' \emph{arXiv preprint arXiv:2311.02962}, 2023.

\bibitem{bi2024codekgc}
Z.~Bi, J.~Chen, Y.~Jiang, F.~Xiong, W.~Guo, H.~Chen, and N.~Zhang, ``Codekgc: Code language model for generative knowledge graph construction,'' \emph{ACM Transactions on Asian and Low-Resource Language Information Processing}, vol.~23, no.~3, pp. 1--16, 2024.

\bibitem{li2023codeie}
P.~Li, T.~Sun, Q.~Tang, H.~Yan, Y.~Wu, X.-J. Huang, and X.~Qiu, ``Codeie: Large code generation models are better few-shot information extractors,'' in \emph{Proceedings of the 61st Annual Meeting of the Association for Computational Linguistics (Volume 1: Long Papers)}, 2023, pp. 15\,339--15\,353.

\bibitem{Lora}
E.~J. Hu, Y.~Shen, P.~Wallis, Z.~Allen-Zhu, Y.~Li, S.~Wang, L.~Wang, and W.~Chen, ``Lora: Low-rank adaptation of large language models,'' \emph{arXiv preprint arXiv:2106.09685}, 2021.

\bibitem{xu2023_S2ynRE}
B.~Xu, Q.~Wang, Y.~Lyu, D.~Dai, Y.~Zhang, and Z.~Mao, ``S2ynre: Two-stage self-training with synthetic data for low-resource relation extraction,'' in \emph{Proceedings of the 61st Annual Meeting of the Association for Computational Linguistics (Volume 1: Long Papers)}, 2023, pp. 8186--8207.

\bibitem{ding2023gptDataAnnotator}
B.~Ding, C.~Qin, L.~Liu, Y.~K. Chia, B.~Li, S.~Joty, and L.~Bing, ``Is gpt-3 a good data annotator?'' in \emph{Proceedings of the 61st Annual Meeting of the Association for Computational Linguistics (Volume 1: Long Papers)}, 2023, pp. 11\,173--11\,195.

\bibitem{chia2022relationprompt}
Y.~K. Chia, L.~Bing, S.~Poria, and L.~Si, ``Relationprompt: Leveraging prompts to generate synthetic data for zero-shot relation triplet extraction,'' \emph{arXiv preprint arXiv:2203.09101}, 2022.

\bibitem{LLM_based_Augmentation_2024}
Z.~Meng, T.~Liu, H.~Zhang, K.~Feng, and P.~Zhao, ``{CEAN}: Contrastive event aggregation network with {LLM}-based augmentation for event extraction,'' in \emph{Proceedings of the 18th Conference of the European Chapter of the Association for Computational Linguistics (Volume 1: Long Papers)}, 2024, pp. 321--333.

\bibitem{synthetic_data1}
Q.~Sun, K.~Huang, X.~Yang, R.~Tong, K.~Zhang, and S.~Poria, ``Consistency guided knowledge retrieval and denoising in llms for zero-shot document-level relation triplet extraction,'' in \emph{Proceedings of the ACM on Web Conference 2024}, ser. WWW '24, 2024, pp. 4407--4416.

\bibitem{synthetic_data2}
J.~Li, Z.~Jia, and Z.~Zheng, ``Semi-automatic data enhancement for document-level relation extraction with distant supervision from large language models,'' in \emph{Proceedings of the 2023 Conference on Empirical Methods in Natural Language Processing (EMNLP)}, 2023, pp. 5495--5505.

\bibitem{zhanghallucination}
Y.~Zhang, Y.~Li, L.~Cui, D.~Cai, L.~Liu, T.~Fu, X.~Huang, E.~Zhao, Y.~Zhang, Y.~Chen \emph{et~al.}, ``Siren's song in the ai ocean: a survey on hallucination in large language models,'' \emph{arXiv preprint arXiv:2309.01219}, 2023.

\bibitem{outdate}
H.~He, H.~Zhang, and D.~Roth, ``Rethinking with retrieval: Faithful large language model inference,'' \emph{arXiv preprint arXiv:2301.00303}, 2022.

\bibitem{domainknowledge}
N.~Kandpal, H.~Deng, A.~Roberts, E.~Wallace, and C.~Raffel, ``Large language models struggle to learn long-tail knowledge,'' in \emph{Proceedings of the 40th International Conference on Machine Learning}, vol. 202, 2023, pp. 15\,696--15\,707.

\bibitem{umls}
O.~Bodenreider, ``The unified medical language system (umls): integrating biomedical terminology,'' \emph{Nucleic acids research}, vol.~32, no. suppl\_1, pp. D267--D270, 2004.

\bibitem{efeoglu2024retrieval_augmented}
S.~Efeoglu and A.~Paschke, ``Retrieval-augmented generation-based relation extraction,'' \emph{arXiv preprint arXiv:2404.13397}, 2024.

\bibitem{denseRetrieval}
V.~Karpukhin, B.~Oguz, S.~Min, P.~Lewis, L.~Wu, S.~Edunov, D.~Chen, and W.-t. Yih, ``Dense passage retrieval for open-domain question answering,'' in \emph{Proceedings of the 2020 Conference on Empirical Methods in Natural Language Processing (EMNLP)}, 2020, pp. 6769--6781.

\bibitem{2024-Retrieval-AugmentedBlack-Box_Language_Models}
W.~Shi, S.~Min, M.~Yasunaga, M.~Seo, R.~James, M.~Lewis, L.~Zettlemoyer, and W.-t. Yih, ``{REPLUG}: Retrieval-augmented black-box language models,'' in \emph{Proceedings of the 2024 Conference of the North American Chapter of the Association for Computational Linguistics: Human Language Technologies (Volume 1: Long Papers)}, 2024, pp. 8364--8377.

\bibitem{zhu2024fcdsfusingconstituencydependency}
X.~Zhu, Z.~Kang, and B.~Hui, ``{FCDS}: Fusing constituency and dependency syntax into document-level relation extraction,'' in \emph{Proceedings of the 2024 Joint International Conference on Computational Linguistics, Language Resources and Evaluation (LREC-COLING 2024)}, 2024, pp. 7141--7152.

\bibitem{DocRE_PSD}
Q.~Wang, Z.~Mao, J.~Gao, and Y.~Zhang, ``Document-level relation extraction with progressive self-distillation,'' \emph{ACM Transactions on Information Systems}, vol.~42, no.~6, 2024.

\bibitem{günther2023jinaEmbeddings}
M.~G{\"u}nther, J.~Ong, I.~Mohr, A.~Abdessalem, T.~Abel, M.~K. Akram, S.~Guzman, G.~Mastrapas, S.~Sturua, B.~Wang \emph{et~al.}, ``Jina embeddings 2: 8192-token general-purpose text embeddings for long documents,'' \emph{arXiv preprint arXiv:2310.19923}, 2023.

\bibitem{BertGT}
P.-T. Lai and Z.~Lu, ``Bert-gt: cross-sentence n-ary relation extraction with bert and graph transformer,'' \emph{Bioinformatics}, vol.~36, no.~24, pp. 5678--5685, 2020.

\bibitem{pubmedBert}
Y.~Gu, R.~Tinn, H.~Cheng, M.~Lucas, N.~Usuyama, X.~Liu, T.~Naumann, J.~Gao, and H.~Poon, ``Domain-specific language model pretraining for biomedical natural language processing,'' \emph{ACM Transactions on Computing for Healthcare (HEALTH)}, vol.~3, no.~1, pp. 1--23, 2021.

\end{thebibliography}
\begin{IEEEbiography}[{\includegraphics[width=1in,height=1.25in,clip,keepaspectratio]{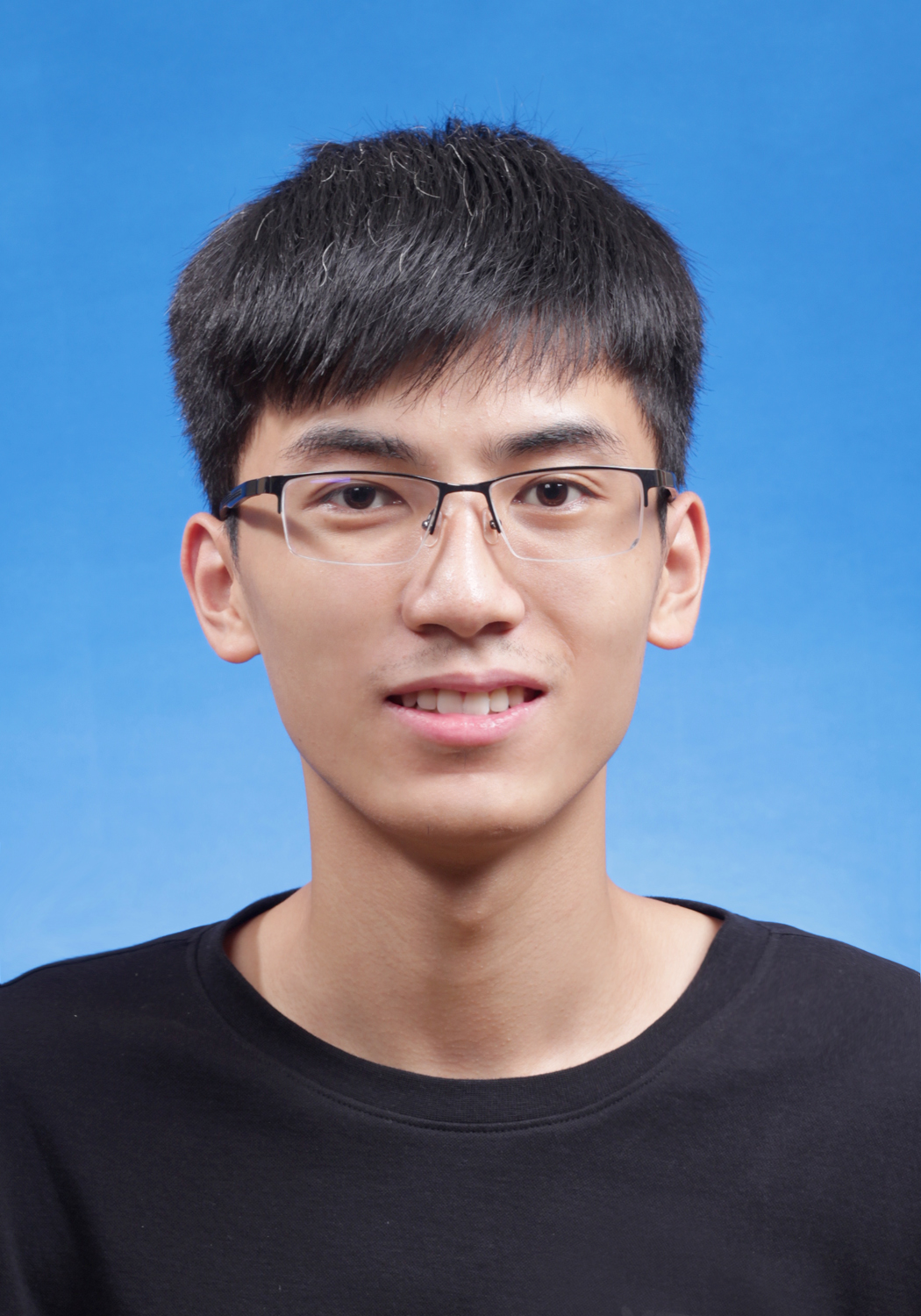}}]{Yufei Shang}
Yufei Shang received his B.E. from Hefei University of Technology, Hefei, in 2023. Now he is a master student at School of Computer Science and Information Engineering, Hefei University of Technology (HFUT). His current research interest is Natural Language Processing.
\end{IEEEbiography}
\begin{IEEEbiography}[{\includegraphics[width=1in,height=1.25in,clip,keepaspectratio]{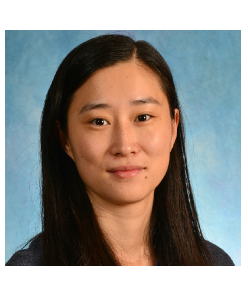}}]{Yanrong Guo}
Yanrong Guo is a professor at School of Computer and Information, Hefei University of Technology (HFUT). She is with Key Laboratory of Knowledge Engineering with Big Data (Hefei University of technology), Ministry of Education. She received her Ph.D. degree at HFUT in 2013. She was a postdoc researcher at University of North Carolina at Chapel Hill (UNC) from 2013 to 2016. Her research interests include pattern recognition and medical image analysis.
\end{IEEEbiography}

\begin{IEEEbiography}[{\includegraphics[width=1in,height=1.25in,clip,keepaspectratio]{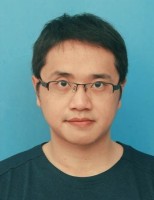}}]{Shijie Hao}
Shijie Hao is a professor at School of Computer Science and Information Engineering, Hefei University of Technology (HFUT). He is also with Key Laboratory of Knowledge Engineering with Big Data (Hefei University of technology), Ministry of Education. He received his Ph.D. degree at HFUT in 2012. His research interests include image processing and pattern recognition.
\end{IEEEbiography}

\begin{IEEEbiography}[{\includegraphics[width=1in,height=1.25in,clip,keepaspectratio]{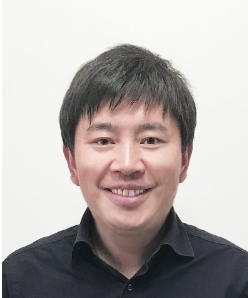}}]{RiChang Hong}
Richang Hong received the Ph.D. degree from the University of Science and Technology of China, Hefei, China, in 2008. He was a Research Fellow of the School of Computing with the National University of Singapore, from 2008 to 2010. He is currently a Professor with the Hefei University of Technology, Hefei. He is also with Key Laboratory of Knowledge Engineering with Big Data (Hefei University of technology), Ministry of Education. He has coauthored over 70 publications in the areas of his research interests, which include multimedia content analysis and social media. He is a member of the ACM and the Executive Committee Member of the ACM SIGMM China Chapter. He was a recipient of the Best Paper Award from the ACM Multimedia 2010, the Best Paper Award from the ACM ICMR 2015, and the Honorable Mention of the IEEE Transactions on Multimedia Best Paper Award. He has served as the Technical Program Chair of the MMM 2016. He has served as an Associate Editor of IEEE Multimedia Magazine, Neural Processing Letter (Springer) Information Sciences (Elsevier) and Signal Processing (Elsevier).
\end{IEEEbiography}

\end{document}